\documentclass[lettersize,journal]{IEEEtran}
\usepackage{amsmath,amsfonts}
\usepackage{algorithmic}
\usepackage{algorithm}
\usepackage{array}
\usepackage[caption=false,font=normalsize,labelfont=sf,textfont=sf]{subfig}
\usepackage{textcomp}
\usepackage{stfloats}
\usepackage{url}
\usepackage{verbatim}
\usepackage{graphicx}
\usepackage{multirow}
\usepackage{caption}
\usepackage{cite}
\usepackage{tabularx}
\usepackage{booktabs}
\usepackage{makecell}
\usepackage[table,dvipsnames]{xcolor}
\usepackage{arydshln}
\usepackage{float}
\usepackage{enumitem}

\begin{document}

\title{FDIO: Frequency Decomposed Inertial Odometry}

\author{
Shanshan Zhang$^{1,2}$, Liqin Wu$^{1}$, Wenying Cao$^{1}$, Lingxiang Zheng$^{1}$, Yu Yang$^{2}$
\thanks{$^{1}$Shanshan Zhang, Liqin Wu, Wenying Cao, and Lingxiang Zheng are with the Department of Information and Communication Engineering, National and Local Joint Engineering Research Center of Navigation and Location Based Services, Xiamen University, Xiamen 361005, China.}
\thanks{$^{2}$Yu Yang and Shanshan Zhang are with the Department of Electronic Science, State Key Laboratory of Physical Chemistry of Solid Surfaces, Xiamen University, Xiamen 361005, China.}
}

\maketitle

\begin{abstract}
Pedestrian inertial odometry (PIO) estimates autonomous pedestrian motion using only acceleration and angular velocity measurements collected by an inertial measurement unit (IMU), making it highly valuable for consumer level localization applications. However, under a dual device acquisition setting, IMU signals collected by a freely carried mobile device are inherently composite signals in which the global motion of the human torso is coupled with perturbations induced by local limb motion. This coupling makes accurate human motion modeling more challenging. To address this issue, this paper proposes frequency decomposed inertial odometry (FDIO). The proposed method first decomposes input IMU signals into low frequency and high frequency components using a Laplacian pyramid. It then adopts a Mamba module to model long range motion information from the low frequency component and uses a multi scale convolution module to extract fine grained local dynamic features from the high frequency component. Experiments on five public PIO datasets show that FDIO achieves an average absolute trajectory error of 3.221~m and an average relative trajectory error of 2.550~m, reducing the errors by 33.3\% and 16.7\% compared with the RoNIN ResNet baseline, respectively. These results validate the effectiveness of the proposed frequency decomposition strategy. To the best of our knowledge, this work is among the first efforts to introduce Mamba and a frequency decomposition architecture into inertial odometry.
\end{abstract}


\section{Introduction}

Inertial measurement units (IMUs) are low cost sensors that operate without external infrastructure and enable privacy preserving sensing. They have therefore been widely integrated into consumer level devices, such as smartphones and wristbands. Pedestrian inertial odometry (PIO) estimates the trajectory of a carrier using only acceleration and angular velocity measurements collected by an IMU. It is valuable for pedestrian navigation, wearable computing, and augmented reality, especially in environments where external information, such as the Global Positioning System (GPS), wireless signals, or visual landmarks, is unavailable or unreliable.

Traditional PIO methods are usually based on Newtonian mechanics and recursively estimate position by numerically integrating IMU measurements. Although these methods have clear physical interpretability, they are highly sensitive to sensor noise and bias. Physical prior constraints can suppress error accumulation under controlled conditions, but such methods usually rely on strong assumptions about sensor placement and motion patterns. In diverse and unconstrained consumer level mobile device scenarios, these assumptions are often difficult to satisfy.

\begin{figure}[t]
\centering
\captionsetup{aboveskip=2pt,font=small}
\includegraphics[width=0.45\textwidth]{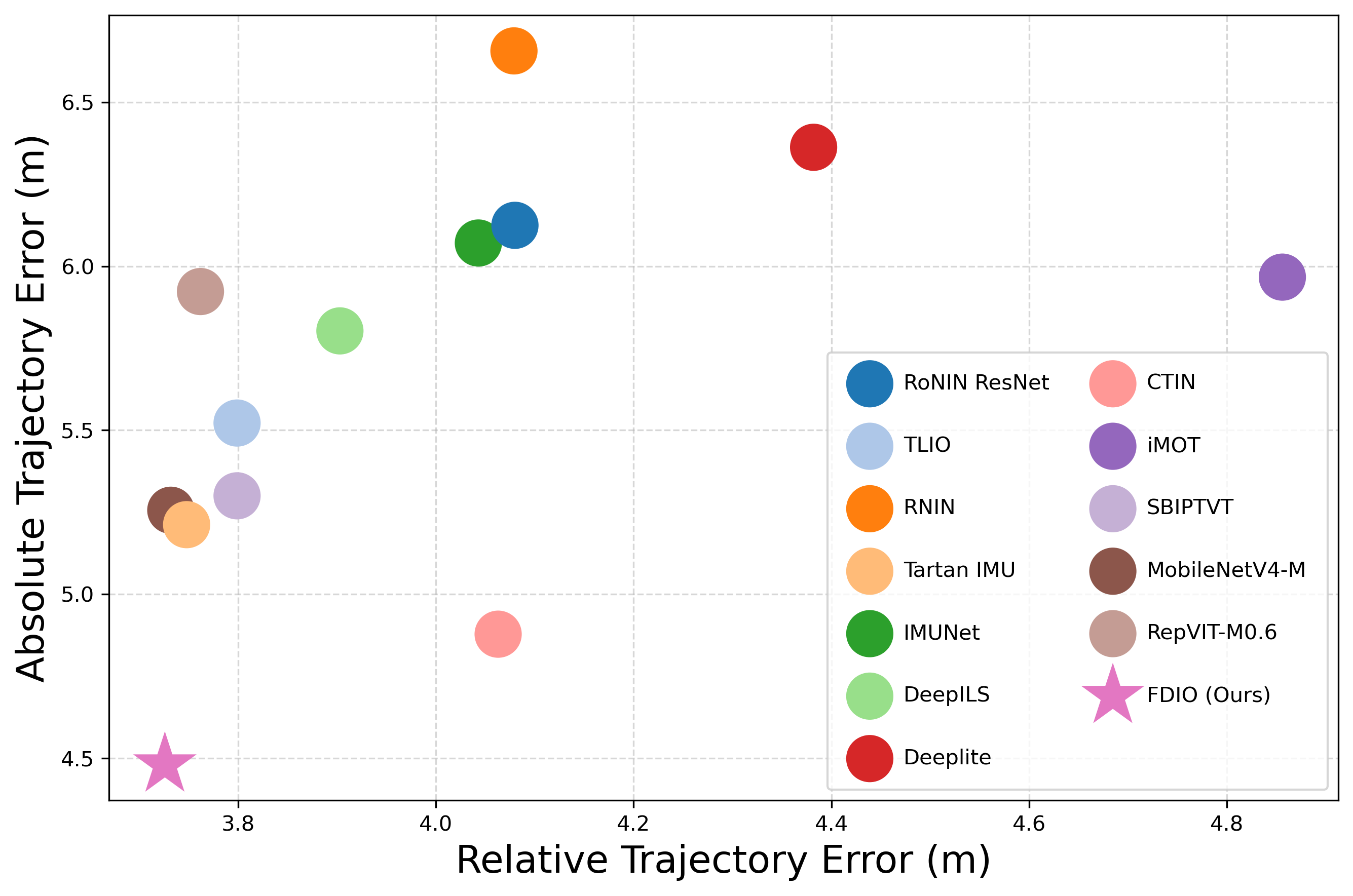}
\caption{Performance comparison of algorithms on the RoNIN dataset in terms of absolute trajectory error (ATE) and relative trajectory error (RTE). Points closer to the lower left corner indicate lower errors and thus higher localization accuracy.}
\label{fig:ATE_vs_RTE}
\vspace{-15pt}
\end{figure}

In recent years, learning based PIO methods have become an important research paradigm. By training neural networks on large scale IMU datasets in an end to end manner, these methods can learn implicit motion patterns that are difficult to characterize explicitly using traditional physical models, thereby improving the estimation accuracy and scenario adaptability of PIO.

However, in pedestrian inertial datasets, IMU signals collected by a freely carried mobile device under a dual device acquisition setting are not clean observations of human torso motion. Instead, they are composite superpositions of multiple motion components. Target motion information dominated by human torso movement usually appears as a low frequency and smoothly varying trend, whereas nontarget motion information, such as local limb perturbations, is more often manifested as broadband high frequency fluctuations. Existing learning based PIO methods typically feed such mixed IMU signals directly into a unified encoder and map them into the same feature space for learning. The coupling among different motion components weakens the ability of the model to distinguish target torso motion from local limb interference, thereby posing a challenge to accurate human motion modeling.

To address this issue, this paper proposes frequency decomposed inertial odometry (FDIO). The proposed method first performs explicit frequency decomposition on IMU signals and then conducts differentiated modeling according to the motion characteristics of different frequency components. This design enhances the capability of the model to represent target motion information. The main contributions of this paper are summarized as follows:

\begin{itemize}[noitemsep,nolistsep,leftmargin=*]
\item We propose a frequency decomposition learning framework for PIO. By explicitly decomposing IMU signals in the frequency domain, the proposed framework constructs differentiated learning paths for motion components with different frequency characteristics.
\item We design a complementary dual path architecture. In this architecture, a Mamba module is used to model long range low frequency motion information, while a multi scale convolution (MSC) module is used to capture high frequency local dynamic features.
\item We conduct systematic experimental evaluations on five public datasets. The results demonstrate that FDIO achieves better localization performance than existing methods, as illustrated in Fig.~\ref{fig:ATE_vs_RTE}.
\end{itemize}

\section{Related Work}

From a methodological perspective, PIO research can be broadly divided into two categories: traditional methods based on Newtonian mechanics and data driven approaches based on machine learning\cite{SurveyofIndoorInertial}. The former relies on explicit physical models and motion priors, whereas the latter learns motion mappings from large scale IMU data to improve localization performance in complex scenarios.

\subsection{Newtonian Mechanics Based Methods}

Early PIO methods, represented by strapdown inertial navigation systems (SINS), estimate the position and orientation of a carrier by numerically integrating acceleration and angular velocity outputs from IMUs\cite{SINS}. However, SINS methods are highly sensitive to measurement accuracy. Even small sensor noise and bias accumulate continuously during integration, eventually leading to significant drift errors\cite{EqNIO}. To mitigate error accumulation, researchers have introduced various physical priors as corrective constraints. For example, pedestrian dead reckoning (PDR) usually assumes a relatively stable gait cycle\cite{PDR}, while zero velocity update (ZUPT) exploits the prior that the velocity of a foot mounted IMU is approximately zero during stance phases for error correction\cite{AdaptiveThreshold-BasedZUPT}. In addition, multisensor fusion methods have been widely adopted to improve system robustness by incorporating auxiliary information, such as gait phase detection\cite{PDR}, vision\cite{A-visual-inertial-approach-to-human-gait-estimation}, or light detection and ranging (LiDAR)\cite{FAST-LIVO2}, and by using extended Kalman filters (EKFs)\cite{RNIN-VIO,MSCKF} for joint state estimation. Although these methods have strong physical interpretability, they usually rely on fixed sensor mounting, stable motion patterns, or additional hardware support. Therefore, their generalization capability and deployment flexibility are limited in freely carried mobile device scenarios and complex real world environments\cite{RoNIN}. More importantly, these methods mainly focus on suppressing integration drift and modeling physical constraints, but they do not explicitly address the coupling between target torso motion and local limb perturbations in freely carried IMU signals.

\subsection{Learning Based Methods}

With the development of deep learning, researchers have explored end to end models that directly learn mappings from inertial measurements to pedestrian velocity or displacement using large scale IMU data, thereby improving the accuracy, robustness, and cross scenario adaptability of PIO\cite{surveyILS}. Early representative works, such as RIDI\cite{RIDI} and PDRNet\cite{PDRNet}, usually first classify the mounting position of the IMU and then regress pedestrian velocity or displacement. Subsequently, RoNIN identified that the device frame used by RIDI is vulnerable to changes in device orientation and proposed the heading aligned coordinate frame (HACF), whose $z$ axis is aligned with gravity, to alleviate representational discontinuities caused by device orientation changes\cite{RoNIN}. RoNIN also systematically compared multiple network architectures, including the temporal convolutional network (TCN)\cite{TCN}, long short term memory (LSTM)\cite{LSTM}, and residual network (ResNet)\cite{ResNet}. Following this work, most subsequent methods adopted HACF as an IMU preprocessing representation and designed convolutional neural networks (CNNs), LSTMs, Transformers, or hybrid architectures on this basis. For example, TLIO\cite{TLIO} and LIDR\cite{LIDR} extended the RoNIN framework by integrating stochastic cloning extended Kalman filters (SCEKFs) and left invariant extended Kalman filters (LIEKFs), respectively. IMUNet\cite{IMUNet}, DeepILS\cite{DeepILS}, and DeepLite\cite{Deep-Learning-for-Inertial-Positioning} mainly focused on lightweight network design. CTIN\cite{CTIN}, SBIPTVT\cite{SBIPTVT}, and iMOT\cite{iMOT} introduced attention mechanisms to enhance temporal modeling. RNIN VIO\cite{RNIN-VIO}, SCHNN\cite{SCHNN}, and SSHNN\cite{SSHNN} constructed multibranch hybrid networks combining CNNs, LSTMs, and attention modules.

Although the above learning based methods have significantly improved the performance of PIO, most of them still treat IMU signals collected by freely carried devices as unified inputs and directly feed them into a single encoder or feature space for learning. This paradigm implicitly assumes that the network can automatically distinguish different motion components from mixed signals. However, in practical free carrying scenarios, IMU signals often contain low frequency and smooth human torso motion trends as well as high frequency dynamic variations caused by arm swing, device relative motion, and local perturbations. These components have distinct frequency distributions and motion semantics. Without an explicit frequency decomposition mechanism, the model may mix target motion information and nontarget perturbation information in the same representation space, thereby weakening its ability to model key motion patterns.

\section{Frequency Characteristics}

Before introducing the FDIO architecture, this section analyzes the frequency characteristics of mobile device IMU signals under the dual device acquisition setting. The purpose of this analysis is to provide empirical motivation for the proposed frequency decomposition architecture.

\begin{figure}[!t]
    \centering
    \includegraphics[width=0.4\textwidth]{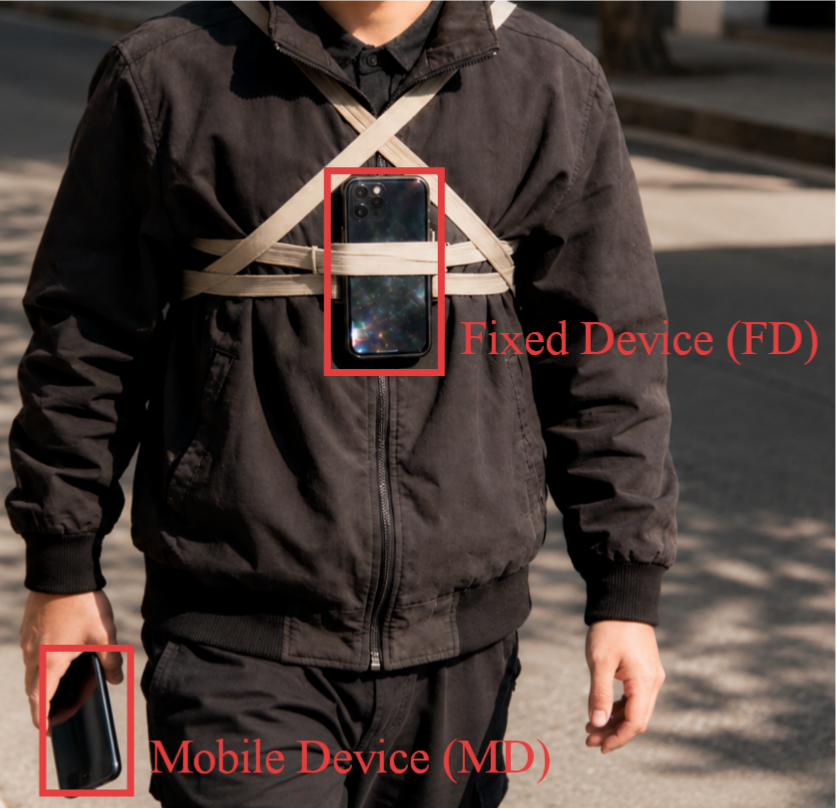}
    \caption{Dual device data acquisition setting, where one device is fixed on the chest as the fixed device (FD) and the other is freely carried as the mobile device (MD).}
    \label{fig:dual_device_acquisition}
    \vspace{-15pt}
\end{figure}

This paper adopts the RoNIN dual device data acquisition setting shown in Fig.~\ref{fig:dual_device_acquisition}. One device is fixed on the chest of the subject and is denoted as the fixed device (FD), which is used to collect IMU signals that are closer to the motion of the human torso. The other device is freely carried by the subject and is denoted as the mobile device (MD), which is used to simulate natural smartphone usage scenarios. The two devices synchronously record IMU measurements during the same walking process. In general, FD signals mainly reflect torso motion, whereas MD signals contain not only torso related motion information but also additional local perturbations introduced by free carrying.

Under ideal conditions where measurement noise, sensor bias, and device attitude estimation errors are ignored, an IMU signal channel measured by the MD can be approximately expressed as the superposition of the corresponding torso related component and the local perturbation component:
\begin{equation}
\mathbf{X}_{\mathrm{MD}}(t) =
\mathbf{X}_{\mathrm{FD}}(t)
+
\mathbf{X}_{\mathrm{LOC}}(t).
\label{eq:imu_signal_model}
\end{equation}
Here, $\mathbf{X}_{\mathrm{MD}}(t)$ denotes the observed IMU signal of the MD, $\mathbf{X}_{\mathrm{FD}}(t)$ denotes the component associated with torso motion, and $\mathbf{X}_{\mathrm{LOC}}(t)$ denotes the local motion component introduced by the free motion of the MD relative to the torso. The signal $\mathbf{X}(t)$ can correspond to different IMU channels, such as acceleration or angular velocity. It should be noted that Eq.~\eqref{eq:imu_signal_model} is not a strict analytical decomposition based on rigid body kinematics. Instead, it serves as an approximate representation for describing the mixture relationship between torso motion relevant to the task and local perturbations in mobile device IMU signals.

\begin{figure}[!t]
    \centering
    \includegraphics[width=0.5\textwidth]{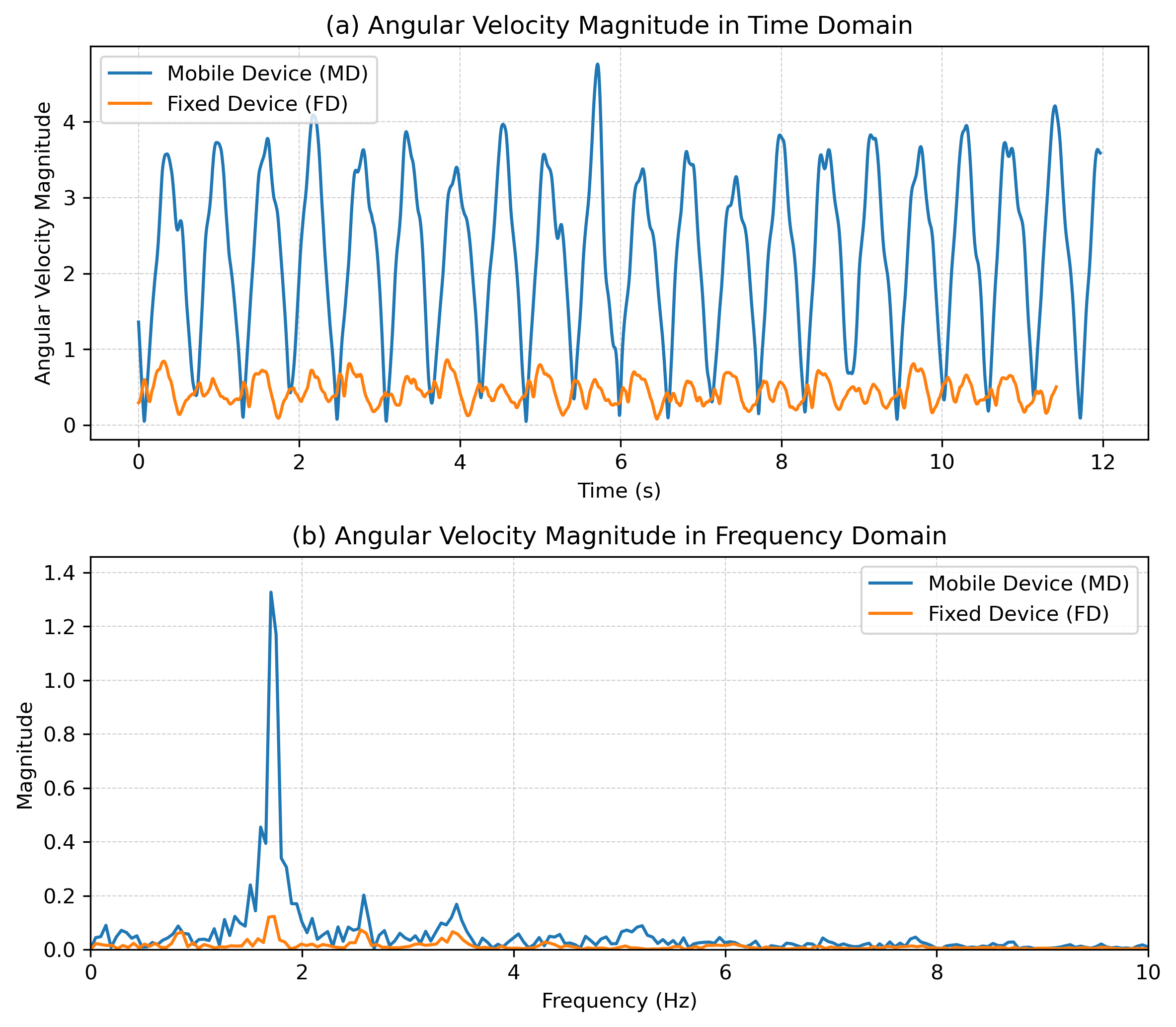}
    \caption{Time domain and frequency domain comparison of angular velocity magnitude collected by the MD and the FD.}
    \label{fig:freq_analysis}
    \vspace{-15pt}
\end{figure}

Fig.~\ref{fig:freq_analysis} compares the angular velocity magnitude signals collected by the FD and the MD under real measurement conditions in both the time domain and the frequency domain. In the time domain, the FD signal remains relatively smooth and has a smaller magnitude, indicating that the fixed device mainly captures stable torso motion. In contrast, the MD signal exhibits larger amplitude variations and more pronounced oscillations, suggesting that the freely carried device is affected by additional local motion.

In the frequency domain, two notable observations can be made. First, both FD and MD signals exhibit a dominant low frequency component near the gait related frequency range, indicating that the main torso motion information is concentrated in the low frequency band. Second, the spectrum of the MD signal is clearly broader than that of the FD signal. It contains multiple secondary peaks and shows stronger energy in higher frequency ranges. This additional high frequency energy mainly originates from local limb motion and device relative motion introduced during free carrying. Similar frequency domain differences can also be observed in acceleration signals, but they are omitted here due to space limitations.

These observations reveal an important structural property of mobile IMU signals. Such signals are frequency heterogeneous mixed signals, in which torso motion related to the localization target and nontarget local perturbations exhibit different frequency distributions. Therefore, exploiting frequency differences can help highlight motion information relevant to the task while suppressing the interference caused by local perturbations during model learning.

\section{Methodology}
\label{sec:methodology}

This section introduces the proposed FDIO. Specifically, we first present the overall architecture of FDIO. We then introduce the frequency decomposed encoder (FDE), which serves as the basic feature extraction unit. The FDE mainly consists of three components: the Laplacian pyramid (LP), MSC, and the Mamba module.

\subsection{Overall Pipeline}
\label{subsec:overall_pipeline}

Given a 1~s IMU observation window, the input sequence can be represented as
$
\mathbf{X}\in\mathbb{R}^{B\times C_{\mathrm{in}}\times T},
$
where $B$ denotes the batch size, $C_{\mathrm{in}}=6$ denotes the six input channels consisting of acceleration along three axes and angular velocity along three axes, and $T=200$ denotes the number of samples within the observation window. The objective of this paper is to learn a mapping from the input IMU sequence to the average velocity within the unit time interval.

\begin{figure*}[!ht]
\centering
\captionsetup{aboveskip=2pt,font=small}
\includegraphics[width=0.95\textwidth]{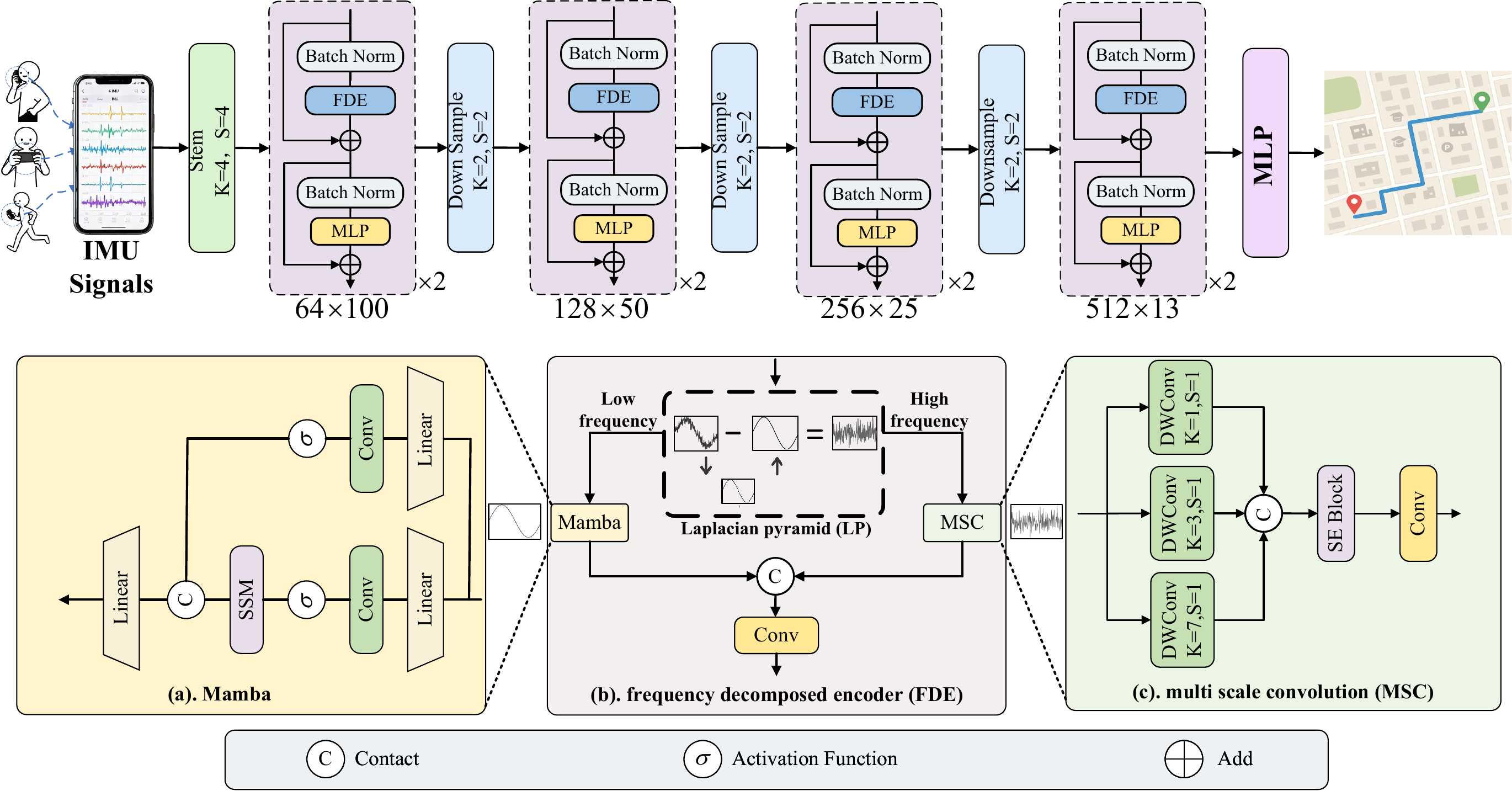}
\caption{Schematic diagram of the proposed FDIO architecture, including the LP decomposition, the MSC module, and the Mamba low frequency modeling branch.}
\label{fig:Framework}
\vspace{-15pt}
\end{figure*}

As shown in Fig.~\ref{fig:Framework}, FDIO adopts a hierarchical temporal feature extraction backbone. First, the input IMU sequence is mapped into a high dimensional feature space through a shallow projection layer, producing the initial feature $\mathbf{X}_{0}$. Then, several FDEs are stacked across multiple levels to progressively extract motion features under different temporal resolutions and channel dimensions. Different from methods that directly encode raw IMU signals in a unified manner, each FDE first performs explicit frequency decomposition on the input feature and then uses different modules to model high frequency local dynamics and low frequency global trends. Through such hierarchical stacking, shallow encoders focus on capturing the basic frequency structure in the raw signal, while deeper encoders further extract more abstract motion patterns.

During training, the mean squared error is used to constrain the difference between the predicted velocity and the ground truth velocity:
\begin{equation}
\mathcal{L}_{\mathrm{vel}}=
\frac{1}{B}\sum_{i=1}^{B}
\left\|
\hat{\mathbf{v}}_{i}-\mathbf{v}_{i}
\right\|_{2}^{2}.
\end{equation}
Here, $\hat{\mathbf{v}}_{i}$ denotes the predicted velocity, and $\mathbf{v}_{i}$ denotes the corresponding ground truth velocity. During inference, the model first predicts the average velocity for consecutive IMU windows and then reconstructs the trajectory through temporal integration.

\subsection{Frequency Decomposed Encoder}
\label{subsec:fde}

Existing learning based PIO methods usually feed IMU signals collected by freely carried devices directly into a unified encoder and learn motion representations in the same feature space. However, such IMU signals are not single component motion signals. Instead, they are composite signals consisting of low frequency human torso motion trends and high frequency local limb perturbations. Without explicitly distinguishing different frequency components, the network has to learn both target motion information and nontarget perturbation information from mixed representations, which increases the difficulty of human motion modeling.

To address this issue, this paper proposes the FDE as the basic encoding unit of FDIO. Given an input feature $\mathbf{X}\in\mathbb{R}^{B\times C\times T}$, the FDE first decomposes it into a low frequency component $\mathbf{X}_{\mathrm{L}}$ and a high frequency component $\mathbf{X}_{\mathrm{H}}$ through the LP. Then, $\mathbf{X}_{\mathrm{H}}$ is fed into the MSC module to extract fine grained local dynamic features, while $\mathbf{X}_{\mathrm{L}}$ is fed into the Mamba module to model long range motion trends. Finally, the outputs of the two paths are fused to obtain a frequency complementary motion representation. This process can be abstracted as
\begin{equation}
\mathrm{FDE}(\mathbf{X})=
\mathrm{Conv}
\left(
\mathrm{Concat}
\left[
\mathrm{MSC}(\mathbf{X}_{\mathrm{H}}),
\mathrm{Mamba}(\mathbf{X}_{\mathrm{L}})
\right]
\right).
\end{equation}

\subsubsection{Laplacian Pyramid}
\label{subsubsec:lp}

The LP is used to explicitly separate low frequency trends and high frequency details in IMU signals before feature extraction. A classical LP usually performs multi scale decomposition through low pass filtering, downsampling, upsampling, and residual computation. Considering that conventional frequency decomposition tools may introduce additional implementation complexity, this paper adopts a lightweight and fully differentiable one dimensional approximation, allowing it to be directly embedded into the deep network and trained in an end to end manner.

Specifically, given an input feature $\mathbf{X}\in\mathbb{R}^{B\times C\times T}$, a one dimensional depth wise convolution (DWConv) with a fixed averaging kernel is used to approximate low pass filtering, and the stride is set to $s=2$ to perform downsampling simultaneously:
\begin{equation}
\mathbf{X}_{\mathrm{ld}} =
\mathrm{DWConv}_{\mathrm{avg}}(\mathbf{X})
\in\mathbb{R}^{B\times C\times \lfloor T/s\rfloor},
\end{equation}
where $\mathrm{DWConv}_{\mathrm{avg}}(\cdot)$ denotes DWConv with a fixed averaging kernel. The averaging kernel $\boldsymbol{\kappa}_{\mathrm{avg}}\in\mathbb{R}^{k}$ satisfies
\begin{equation}
\sum_{j=1}^{k}\boldsymbol{\kappa}_{\mathrm{avg}}(j)=1.
\end{equation}
In our experiments, $k=5$ is used to achieve a tradeoff between smoothing local fluctuations and preserving the main motion trend. Then, nearest neighbor upsampling is used to restore the low resolution low frequency representation to the original temporal length:
\begin{equation}
\mathbf{X}_{\mathrm{L}} =
\mathrm{Up}(\mathbf{X}_{\mathrm{ld}})
\in\mathbb{R}^{B\times C\times T}.
\end{equation}
The high frequency component is obtained by computing the residual between the original input and the low frequency component:
\begin{equation}
\mathbf{X}_{\mathrm{H}} =
\mathbf{X}-\mathbf{X}_{\mathrm{L}}
\in\mathbb{R}^{B\times C\times T}.
\end{equation}
Through the above decomposition, $\mathbf{X}_{\mathrm{L}}$ mainly preserves smooth variations such as human torso motion trends, while $\mathbf{X}_{\mathrm{H}}$ mainly contains instantaneous changes, local perturbations, and fine grained dynamic patterns.

\subsubsection{Multi Scale Convolution}
\label{subsubsec:msc}

The high frequency component $\mathbf{X}_{\mathrm{H}}$ obtained by the LP contains local dynamics at multiple scales, such as instantaneous gait variations, turning perturbations, and local limb motion caused by freely carried devices. If only a single convolution kernel is used for modeling, the network may have difficulty adapting to high frequency patterns with different durations and scales. Therefore, this paper designs the MSC module, which extracts multi scale local features through multiple parallel DWConv branches.

Specifically, the MSC module contains three convolution branches with kernel sizes of 1, 3, and 7, respectively. For the input high frequency component $\mathbf{X}_{\mathrm{H}}$, the three branches are computed as
\begin{align}
\mathbf{X}_{1} &= \mathrm{DWConv}_{k=1}(\mathbf{X}_{\mathrm{H}}),\\
\mathbf{X}_{3} &= \mathrm{DWConv}_{k=3}(\mathbf{X}_{\mathrm{H}}),\\
\mathbf{X}_{7} &= \mathrm{DWConv}_{k=7}(\mathbf{X}_{\mathrm{H}}).
\end{align}
The branch with $k=1$ emphasizes instantaneous channel responses, the branch with $k=3$ extracts short term local patterns, and the branch with $k=7$ provides a larger local receptive field for modeling dynamic changes with slightly longer durations. Subsequently, the outputs of the three branches are concatenated along the channel dimension, and a squeeze and excitation (SE) block and point wise convolution (PWConv) are used for adaptive reweighting and channel fusion:
\begin{equation}
\mathbf{X}_{\mathrm{MSC}} =
\mathrm{PWConv}
\left(
\mathrm{SE}
\left(
\mathrm{Concat}
\left[
\mathbf{X}_{1},\mathbf{X}_{3},\mathbf{X}_{7}
\right]
\right)
\right),
\end{equation}
where $\mathrm{PWConv}(\cdot)$ denotes PWConv. With this design, the MSC module enhances the ability of the high frequency branch to represent local motion patterns at different scales with low computational complexity.

\subsubsection{Mamba Module}
\label{subsubsec:mamba}

Different from the high frequency component, the low frequency component $\mathbf{X}_{\mathrm{L}}$ mainly reflects the overall human motion trend, and its key patterns are usually distributed over a longer temporal range. Therefore, the low frequency branch requires the ability to model long range dependencies. This paper adopts the Mamba module to model $\mathbf{X}_{\mathrm{L}}$ and extract global temporal information with linear complexity.

Mamba is based on the state space model (SSM). The standard continuous time SSM can be formulated as
\begin{align}
\dot{\mathbf{h}}(t) &= \mathbf{A}_{s}\mathbf{h}(t)+\mathbf{B}_{s}\mathbf{X}(t),\\
\mathbf{y}(t) &= \mathbf{C}_{s}\mathbf{h}(t),
\end{align}
where $\mathbf{X}(t)$ denotes the input, $\mathbf{h}(t)$ denotes the hidden state, $\mathbf{y}(t)$ denotes the output, and $\mathbf{A}_{s}$, $\mathbf{B}_{s}$, and $\mathbf{C}_{s}$ are learnable SSM parameters. After discretization, the SSM can be written as
\begin{align}
\mathbf{h}_{t} &= \bar{\mathbf{A}}_{s}\mathbf{h}_{t-1}+\bar{\mathbf{B}}_{s}\mathbf{X}_{t},\\
\mathbf{y}_{t} &= \mathbf{C}_{s}\mathbf{h}_{t}.
\end{align}
Mamba introduces an input dependent selective mechanism on top of the SSM, enabling the model to dynamically select contextual information that is more relevant to the current motion state according to the input content.

In implementation, this paper adopts the MambaVision structure, which combines a selective SSM branch with a convolution branch. For the low frequency input $\mathbf{X}_{\mathrm{L}}$, the computation can be abstracted as
\begin{align}
\mathbf{Z}_{1}
&=
\mathrm{ScanSSM}
\left(
\mathrm{Act}
\left(
\mathrm{Conv}
\left(
\mathrm{Linear}(\mathbf{X}_{\mathrm{L}})
\right)
\right)
\right),\\
\mathbf{Z}_{2}
&=
\mathrm{Act}
\left(
\mathrm{Conv}
\left(
\mathrm{Linear}(\mathbf{X}_{\mathrm{L}})
\right)
\right),\\
\mathbf{X}_{\mathrm{Mamba}}
&=
\mathrm{Linear}
\left(
\mathrm{Concat}
\left[
\mathbf{Z}_{1},\mathbf{Z}_{2}
\right]
\right).
\end{align}
Here, $\mathrm{ScanSSM}(\cdot)$ denotes selective state space scanning, and $\mathrm{Act}(\cdot)$ denotes a nonlinear activation function. This structure uses the SSM branch to model long range motion trends in low frequency signals and the convolution branch to supplement local context, thereby producing a more stable low frequency motion representation.

Finally, the FDE fuses the high frequency branch output $\mathbf{X}_{\mathrm{MSC}}$ and the low frequency branch output $\mathbf{X}_{\mathrm{Mamba}}$:
\begin{equation}
\mathbf{X}_{\mathrm{out}} =
\mathrm{Conv}
\left(
\mathrm{Concat}
\left[
\mathbf{X}_{\mathrm{MSC}},
\mathbf{X}_{\mathrm{Mamba}}
\right]
\right).
\end{equation}
The output $\mathbf{X}_{\mathrm{out}}$ contains both high frequency local dynamic information and low frequency global trend information, and it is fed into subsequent network layers as the output of the current FDE. Through this decomposition, modeling, and fusion scheme, FDIO can more fully exploit the complementarity of different frequency components, thereby improving the modeling capability of PIO in freely carried device scenarios.

\begin{table*}[htbp]
  \centering
  \small
  \setlength{\tabcolsep}{2pt}
  \setlength{\arrayrulewidth}{0.8pt}
  \renewcommand{\arraystretch}{1.2}
  \setlength{\aboverulesep}{0pt}
  \setlength{\belowrulesep}{0pt}
  \caption{Algorithm performance across datasets. Best and second best results are marked in orange and yellow, respectively.}
  \label{tab:exp_results}
  \begin{tabular}{c|c|cccccccccc|cc}
    \hline
    \multicolumn{2}{c|}{Dataset} & \multicolumn{2}{c}{RIDI} & \multicolumn{2}{c}{RoNIN}  & \multicolumn{2}{c}{RNIN} & \multicolumn{2}{c}{IMUNet} & \multicolumn{2}{c|}{OxIOD} & \multicolumn{2}{c}{Average} \\
    \hline
    \multicolumn{2}{c|}{Metrics in meters} & ATE & RTE & ATE & RTE & ATE & RTE & ATE & RTE & ATE & RTE & ATE & RTE  \\
    \hline
    \multirow{2}{*}{\makecell[c]{Hybrid\\based}}
    & FDIO (Ours) &
    \cellcolor{orange!30}1.750 & \cellcolor{orange!30}1.996 & \cellcolor{orange!30}4.480 & \cellcolor{orange!30}3.726 & \cellcolor{orange!30}3.002 & \cellcolor{yellow!40}2.166 & \cellcolor{orange!30}5.303 & \cellcolor{orange!30}3.882 & \cellcolor{orange!30}1.568 & \cellcolor{green!10}0.980 & \cellcolor{orange!30}3.221 & \cellcolor{orange!30}2.550  \\
    & RNIN (ISMAR 2021)\cite{RNIN-VIO}  & \cellcolor{green!10}2.771 & \cellcolor{green!10}2.992 & \cellcolor{green!10}6.657 & \cellcolor{green!10}4.079 & \cellcolor{green!10}4.203 & \cellcolor{green!10}2.285 & \cellcolor{green!10}8.400 & \cellcolor{green!10}5.055 & \cellcolor{green!10}3.662 & \cellcolor{green!10}4.337 & \cellcolor{green!10}5.139 & \cellcolor{green!10}3.750  \\
    \hline
    \multirow{8}{*}{\makecell[c]{CNN\\based}}
    & RoNIN ResNet (ICRA 2020)\cite{RoNIN} & \cellcolor{green!10}2.429 & \cellcolor{green!10}2.599 & \cellcolor{green!10}6.125 & \cellcolor{green!10}4.080 & \cellcolor{green!10}3.924 & \cellcolor{green!10}2.621 & \cellcolor{green!10}8.159 & \cellcolor{green!10}4.706 & \cellcolor{green!10}3.498 & \cellcolor{green!10}1.305 & \cellcolor{green!10}4.827 & \cellcolor{green!10}3.062 \\
    & TLIO (RA L 2020)\cite{TLIO}   & \cellcolor{green!10}2.240 & \cellcolor{green!10}2.524 & \cellcolor{green!10}5.522 & \cellcolor{green!10}3.799 & \cellcolor{green!10}4.297 & \cellcolor{green!10}3.353 & \cellcolor{green!10}7.490 & \cellcolor{green!10}5.273 & \cellcolor{green!10}2.020 & \cellcolor{green!10}0.969 & \cellcolor{green!10}4.314 & \cellcolor{green!10}3.184 \\
    & IMUNet (TIM 2024)\cite{IMUNet} & \cellcolor{green!10}3.201 & \cellcolor{green!10}3.240 & \cellcolor{green!10}6.071 & \cellcolor{green!10}4.043 & \cellcolor{yellow!40}3.156 & \cellcolor{green!10}2.262 & \cellcolor{green!10}8.676 & \cellcolor{green!10}5.947 & \cellcolor{yellow!40}1.638 & \cellcolor{green!10}1.043 & \cellcolor{green!10}4.548 & \cellcolor{green!10}3.307 \\
    & DeepILS (IoT Journal 2025)\cite{DeepILS}   & \cellcolor{yellow!40}1.893 & \cellcolor{green!10}2.363 & \cellcolor{green!10}5.803 & \cellcolor{green!10}3.903 & \cellcolor{green!10}3.675 & \cellcolor{green!10}2.392 & \cellcolor{green!10}7.861 & \cellcolor{green!10}5.585 & \cellcolor{green!10}3.294 & \cellcolor{green!10}1.233 & \cellcolor{green!10}4.505 & \cellcolor{green!10}3.095 \\
    & DeepLite (IoT Journal 2025)\cite{DeepLite}   & \cellcolor{green!10}3.658 & \cellcolor{green!10}3.814 & \cellcolor{green!10}6.363 & \cellcolor{green!10}4.382 & \cellcolor{green!10}3.878 & \cellcolor{green!10}2.892 & \cellcolor{green!10}7.459 & \cellcolor{green!10}5.746 & \cellcolor{green!10}2.048 & \cellcolor{green!10}1.112 & \cellcolor{green!10}4.681 & \cellcolor{green!10}3.589 \\
    & RepViT M0.6 (CVPR 2024)\cite{Repvit} & \cellcolor{green!10}2.523 & \cellcolor{green!10}2.880 & \cellcolor{green!10}5.923 & \cellcolor{green!10}3.762 & \cellcolor{green!10}3.916 & \cellcolor{green!10}2.503 & \cellcolor{green!10}7.103 & \cellcolor{green!10}5.141 & \cellcolor{green!10}2.859 & \cellcolor{green!10}1.165 & \cellcolor{green!10}4.465 & \cellcolor{green!10}3.090 \\
    & MobileNetV4 S (arXiv 2024)\cite{MobileNetV4} & \cellcolor{green!10}2.677 & \cellcolor{green!10}3.141 & \cellcolor{green!10}5.256 & \cellcolor{yellow!40}3.732 & \cellcolor{green!10}3.768 & \cellcolor{green!10}2.567 & \cellcolor{green!10}7.787 & \cellcolor{green!10}6.661 & \cellcolor{green!10}4.408 & \cellcolor{green!10}1.617 & \cellcolor{green!10}4.779 & \cellcolor{green!10}3.544 \\
    & Tartan IMU (CVPR 2025)\cite{Tartan-IMU}  & \cellcolor{green!10}2.254 & \cellcolor{green!10}2.581 & \cellcolor{green!10}5.212 & \cellcolor{green!10}3.748 & \cellcolor{green!10}3.169 & \cellcolor{orange!30}1.982 & \cellcolor{green!10}6.291 & \cellcolor{green!10}4.492 & \cellcolor{green!10}1.670 & \cellcolor{orange!30}0.940 & \cellcolor{yellow!40}3.719 & \cellcolor{green!10}2.749 \\
    \hline
    \multirow{3}{*}{\makecell[c]{Transformer\\based}}
    & iMOT (AAAI 2025)\cite{iMOT}  & \cellcolor{green!10}2.788 & \cellcolor{green!10}3.485 & \cellcolor{green!10}5.967 & \cellcolor{green!10}4.856 & \cellcolor{green!10}6.359 & \cellcolor{green!10}3.229 & \cellcolor{green!10}8.953 & \cellcolor{green!10}6.616 & \cellcolor{green!10}21.802 & \cellcolor{green!10}5.197 & \cellcolor{green!10}9.174 & \cellcolor{green!10}4.677 \\
    & CTIN (AAAI 2022)\cite{CTIN}  & \cellcolor{green!10}2.325 & \cellcolor{green!10}2.918 & \cellcolor{yellow!40}4.878 & \cellcolor{green!10}4.063 & \cellcolor{green!10}4.732 & \cellcolor{green!10}2.813 & \cellcolor{yellow!40}5.589 & \cellcolor{green!10}4.522 & \cellcolor{green!10}4.846 & \cellcolor{green!10}1.563 & \cellcolor{green!10}4.474 & \cellcolor{green!10}3.176 \\
    & SBIPTVT (CSCWD 2024)\cite{SBIPTVT} & \cellcolor{green!10}2.074 & \cellcolor{yellow!40}2.317 & \cellcolor{green!10}5.300 & \cellcolor{green!10}3.799 & \cellcolor{green!10}3.781 & \cellcolor{green!10}2.269 & \cellcolor{green!10}6.466 & \cellcolor{yellow!40}4.325 & \cellcolor{green!10}1.732 & \cellcolor{yellow!40}0.954 & \cellcolor{green!10}3.871 & \cellcolor{yellow!40}2.733 \\
    \hline
    \multirow{1}{*}{Improvement}
    & FDIO vs RoNIN ResNet(\%)  & \cellcolor{green!10}28.0 & \cellcolor{green!10}23.2 & \cellcolor{green!10}26.9 & \cellcolor{green!10}8.7 & \cellcolor{green!10}23.5 & \cellcolor{green!10}17.4 & \cellcolor{green!10}35.0 & \cellcolor{green!10}17.5 & \cellcolor{green!10}55.2 & \cellcolor{green!10}24.9 & \cellcolor{green!10}33.3 & \cellcolor{green!10}16.7 \\
    \hline
  \end{tabular}
\vspace{-10pt}
\end{table*}

\begin{figure*}[htbp]
    \centering
    \includegraphics[width=0.95\textwidth]{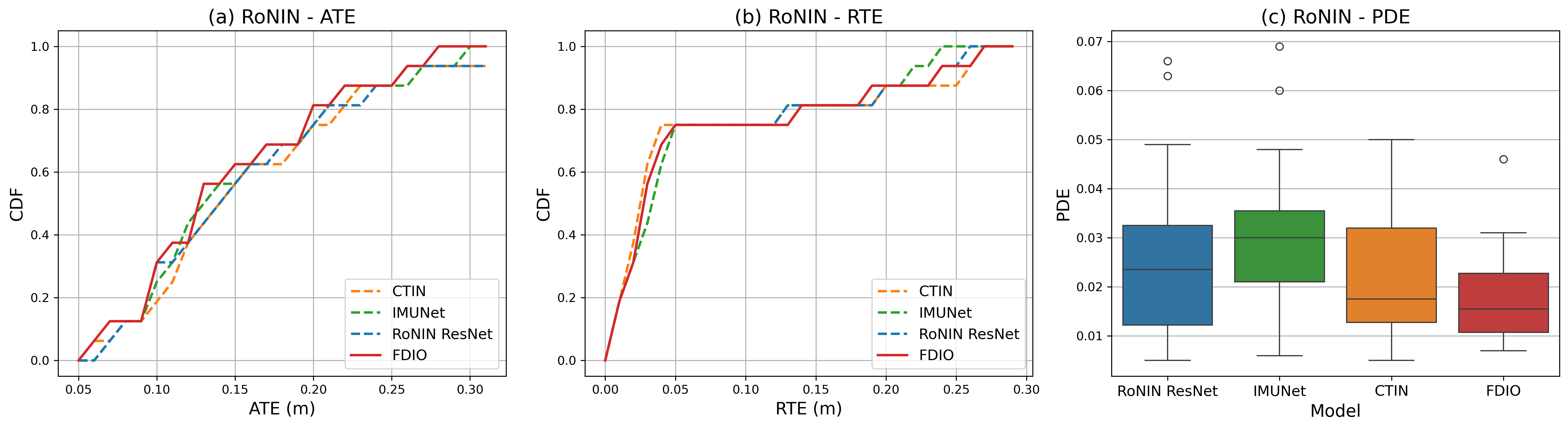}
    \caption{Error distribution analysis on the RoNIN dataset. The cumulative distribution functions of ATE and RTE and the boxplot of position drift error (PDE) are shown for different methods.}
    \label{fig:fdio_cdf_pde}
    \vspace{-15pt}
\end{figure*}

\section{Experiments}
\subsection{Experimental Setup}
\subsubsection{Datasets}

Experiments are conducted on five public pedestrian PIO datasets, including RIDI\cite{RIDI}, RoNIN\cite{RoNIN}, RNIN\cite{RNIN-VIO}, OxIOD\cite{Oxiod}, and IMUNet\cite{IMUNet}. Each dataset is divided into training, testing, and validation sets with a ratio of 8:1:1. These datasets cover diverse indoor and outdoor environments and include various motion patterns, such as walking, running, and stair climbing, providing a comprehensive basis for evaluating model generalization and robustness.

\subsubsection{Baseline Methods}

Considering that learning based approaches usually significantly outperform traditional physics driven methods in localization accuracy, representative deep learning models are selected as baselines. Specifically, the baselines include CNN based methods, such as RoNIN ResNet\cite{RoNIN}, IMUNet\cite{IMUNet}, DeepLite\cite{DeepLite}, DeepILS\cite{DeepILS}, and the network proposed in TLIO\cite{TLIO}; attention based methods, such as CTIN\cite{CTIN}, iMOT\cite{iMOT}, and SBIPTVT\cite{SBIPTVT}; and hybrid architectures, such as the networks proposed in RNIN\cite{RNIN-VIO} and Tartan IMU\cite{Tartan-IMU}. In addition, following the experimental settings of IMUNet and DeepLite, general purpose networks that have shown strong performance in other domains, including RepViT\cite{Repvit} and MobileNetV4\cite{MobileNetV4}, are further included to enable a fair cross architecture comparison. For methods without publicly available code, including CTIN\cite{CTIN}, iMOT\cite{iMOT}, SBIPTVT\cite{SBIPTVT}, and Tartan IMU\cite{Tartan-IMU}, we reproduced their implementations.

\subsubsection{Evaluation Metrics}

Three commonly used evaluation metrics in PIO are adopted: PDE, ATE, and RTE\cite{Ionet}. PDE measures the ratio between the endpoint position drift of the entire trajectory and the total length of the ground truth trajectory. ATE computes the root mean square error (RMSE) between the estimated trajectory and the ground truth trajectory after global alignment, reflecting overall localization accuracy. RTE computes the RMSE within a fixed length sliding window and evaluates short term consistency and drift control.

\subsubsection{Training Details}

The model adopts four hierarchical feature extraction stages with channel dimensions of $[64,128,256,512]$. The network is optimized using the Adam optimizer with an initial learning rate of $10^{-4}$, and training is terminated early when the learning rate decays to $10^{-6}$. The maximum number of training epochs is set to 40. All experiments are conducted on five NVIDIA RTX 3090 GPUs using Compute Unified Device Architecture (CUDA) 12.2 and PyTorch 2.5.1.

\subsection{Algorithm Comparison}

Table~\ref{tab:exp_results} summarizes the quantitative results of all methods on the five public pedestrian PIO datasets. The experimental results show that the proposed FDIO outperforms existing representative methods in most scenarios. For example, on the RoNIN dataset, FDIO reduces ATE by approximately 26.9\% and RTE by approximately 8.7\% compared with RoNIN ResNet. In addition, in terms of the average results across the five datasets, FDIO achieves the best localization accuracy. Compared with RoNIN ResNet, its ATE and RTE are reduced by 33.3\% and 16.7\%, respectively. These results quantitatively validate the effectiveness of the proposed method in improving localization accuracy.

\begin{figure*}[!t]
\centering
\captionsetup{aboveskip=2pt,font=small}
\includegraphics[width=1\textwidth]{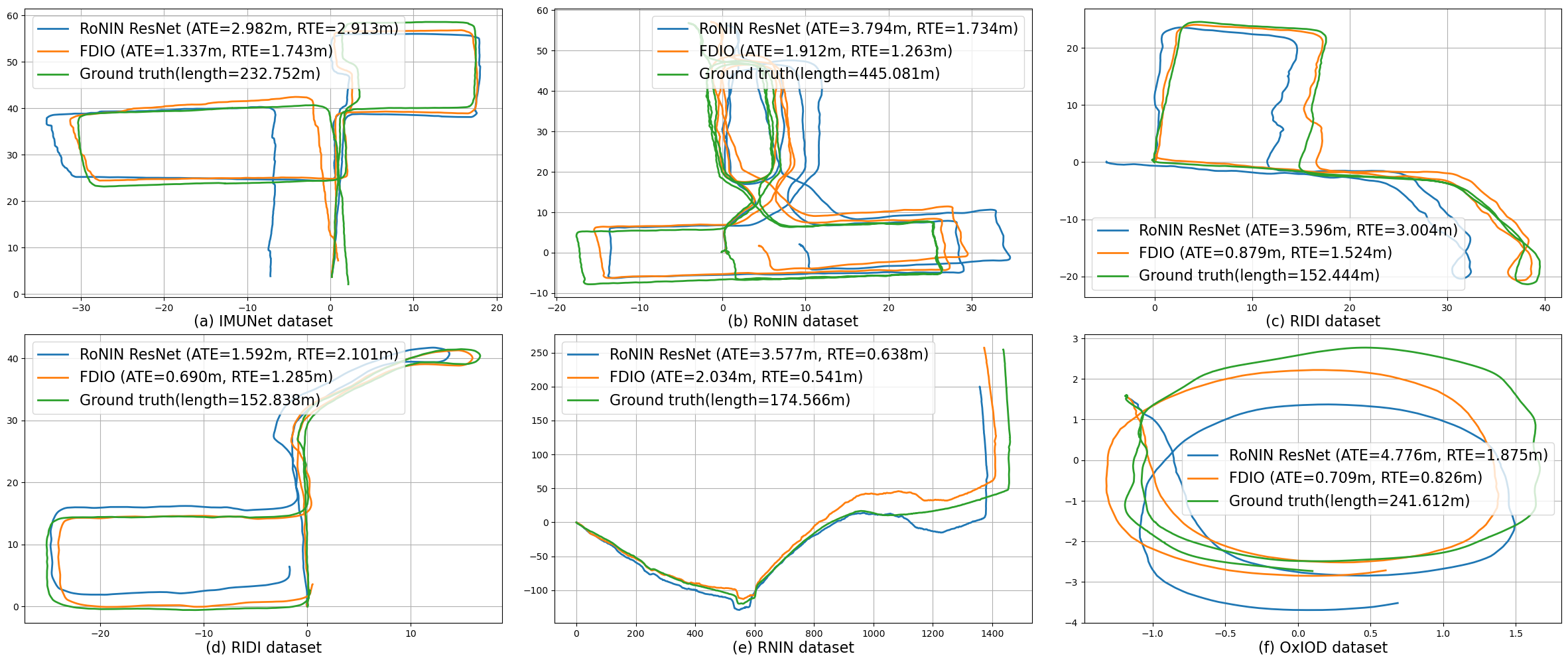}
\caption{Representative trajectory visualization results on five datasets, comparing FDIO with RoNIN ResNet. For better visualization clarity, only the first 10\% of the trajectory is shown for the OxIOD dataset.}
\label{combined_plot}
\end{figure*}

\begin{table}[!tbp]
\centering
\captionsetup{aboveskip=2pt, font=small}
\small
\setlength{\arrayrulewidth}{0.8pt}
\setlength{\aboverulesep}{0pt}
\setlength{\belowrulesep}{0pt}
\caption{Ablation study results. The lowest value in each row is highlighted in bold. All metrics are in meters.}
\label{ablation_study}
\begin{tabular*}{\linewidth}{@{\extracolsep{\fill}} ll|ccc}
\hline
Dataset & Metric
& Conv only
& Mamba only
& FDIO (full)\\
\hline
\multirow{2}{*}{RoNIN}
& ATE & 6.835 & 6.063 & \textbf{4.480} \\
& RTE & 4.160 & 4.000 & \textbf{3.726} \\

\multirow{2}{*}{RIDI}
& ATE & 2.318 & 2.114 & \textbf{1.750} \\
& RTE & 2.656 & 2.419 & \textbf{1.996} \\

\multirow{2}{*}{IMUNet}
& ATE & 6.132 & 6.366 & \textbf{5.303} \\
& RTE & 4.207 & 4.493 & \textbf{3.882} \\

\multirow{2}{*}{RNIN}
& ATE & 6.154 & 3.907 & \textbf{3.002} \\
& RTE & 3.223 & 2.847 & \textbf{2.166} \\

\multirow{2}{*}{OxIOD}
& ATE & 1.844 & 2.723 & \textbf{1.568} \\
& RTE & 1.020 & 1.167 & \textbf{0.980} \\
\hline
\end{tabular*}
\vspace{-15pt}
\end{table}

To further analyze the overall characteristics of the error distributions, the RoNIN dataset is used as an example to visualize the cumulative distribution functions (CDFs) of ATE and RTE as well as the boxplot of PDE, as shown in Fig.~\ref{fig:fdio_cdf_pde}. As shown in Fig.~\ref{fig:fdio_cdf_pde}(a) and Fig.~\ref{fig:fdio_cdf_pde}(b), the curves corresponding to FDIO are closer to the upper left region, indicating that FDIO covers a higher proportion of test samples under the same error threshold. In other words, under the same cumulative probability, FDIO tends to have smaller ATE and RTE. In addition, the PDE boxplot in Fig.~\ref{fig:fdio_cdf_pde}(c) shows that FDIO has a lower median, a more compact box range, and a smaller overall dispersion. In contrast, the PDE distributions of RoNIN ResNet, IMUNet, and CTIN are more scattered and contain more obvious high error samples. These observations demonstrate that FDIO can not only effectively reduce localization errors but also maintain more stable endpoint prediction performance.

To provide a more intuitive evaluation of localization performance, Fig.~\ref{combined_plot} presents representative trajectory results from each dataset. For relatively simple straight or gently curved trajectories, both RoNIN ResNet and FDIO can closely follow the ground truth paths. In contrast, for trajectories with multiple turns or complex dynamics, RoNIN ResNet exhibits more noticeable deviations, whereas FDIO maintains higher trajectory reconstruction accuracy. This improvement is mainly attributed to the multi scale decomposition of IMU signals by the LP, which enables the model to capture both local details and global trends, leading to a more robust and accurate motion representation.

\subsection{Ablation Study}

To evaluate the effectiveness of the proposed components, two ablation variants are constructed: Conv only and Mamba only. The Conv only variant processes raw IMU signals using only the multipath convolution module, while the Mamba only variant processes raw IMU signals using only the Mamba module. Table~\ref{ablation_study} presents the performance of these variants and the full FDIO model across the five datasets.

The experimental results show that although Conv only has basic localization capability, its accuracy is clearly lower than that of the full FDIO, indicating that local modeling alone is insufficient to capture the complex multi scale dynamics in IMU signals. Similarly, Mamba only also shows limited performance, suggesting that purely global modeling cannot adequately represent high frequency local details such as gait transients or turning jitter. In contrast, the complete FDIO achieves the best results on all datasets, further validating the complementary advantages of the high and low frequency decomposition strategy and the dual path collaborative architecture.

\section{Conclusion}

This paper proposes FDIO, a frequency decomposition learning framework for PIO. The proposed method decomposes IMU signals into high frequency and low frequency components using an LP, and it models local details and global motion trends through an MSC module and a Mamba module, respectively. Experimental results show that FDIO achieves better localization accuracy on multiple public datasets, validating the effectiveness of explicit frequency decomposition and dual path collaborative modeling. However, the high frequency component still contains interference noise that is difficult to separate precisely, and existing methods remain limited in further suppressing such noise. Therefore, designing a more effective separation mechanism for target motion and nontarget perturbations will be an important direction for future research.

\bibliographystyle{IEEEtran}
\bibliography{refs}

\end{document}